%% file: paper.tex
\PassOptionsToPackage{table}{xcolor}
\RequirePackage{fix-cm}
\documentclass[]{bytedance_seed}

\usepackage[toc,page,header]{appendix}
\usepackage{minitoc}
\usepackage{amsmath}
\usepackage{amssymb}

\usepackage{url}
\usepackage{tikz}
\usetikzlibrary{arrows.meta,calc}
\usepackage{pifont}
\usepackage{makecell}
\usepackage{float}
\usepackage{wrapfig}
\usepackage{needspace}

\titleformat*{\paragraph}{\bfseries}

\title{HumanSightBench: A Benchmark for Human Visual Perception in MLLMs}

\author[1,3,\dagger]{Zixin~Su}
\author[1,2,*]{Yangguang~Lin}
\author[1,3,*]{Zegang~Peng}
\author[1,4,*]{Yuntong~Sun}
\author[1]{Xuanyi~Wang}
\author[1,2,*]{Jiahao~Wang}
\author[1,5,*]{Qiankai~Xu}
\author[1,3,*]{Bing~Yuan}
\author[1]{Jingjia~Huang}
\author[1]{Ruyang~Liu}
\author[1]{Yanxu~Hu}
\author[1]{Zufeng~Zhou}
\author[1]{Yi~Lin}

\affiliation[1]{ByteDance Seed}
\affiliation[2]{Beijing University of Posts and Telecommunications}
\affiliation[3]{Tsinghua University}
\affiliation[4]{Renmin University of China}
\affiliation[5]{Nanjing University}

\contribution[*]{Work done during an internship at ByteDance Seed}
\contribution[\dagger]{Corresponding author}

\abstract{
Multimodal large language models (MLLMs) have demonstrated strong performance on knowledge-intensive and reasoning-based vision-language benchmark datasets, but still struggle to solve the visual questions that people can solve easily. Evaluating the differences between MLLMs and human-level visual perception is challenging due to end-to-end evaluation which combines various skills, or because we are uncertain if individual items are valid. Specifically, observed errors may reflect visual perception, language priors, background knowledge, question ambiguity, and underspecified answers. Our guideline for HumanSightBench is that we construct items that are human-easy by construction and model-hard by evaluation. HumanSightBench contains 2,876 visual questions with short answers across 7 top-level perceptual categories and 21 subcategories. We only include items that pass five inclusion gates: question clarity, answer uniqueness, visual grounding, human solvability, and text-only shortcut filtering. According to the protocol, errors are defined as incorrect answers on human-validated visual tasks with the evaluated interface. The inclusion gates are intended to reduce confounds from ambiguity and external knowledge. The best pass@3 score is 65.6\% across eight evaluated MLLMs. No evaluated model exceeds 27.5\% on Counting. These results identify capability gaps that are hidden by overall scores and can help future training experiments and held-out evaluation.
}

\date{\today}

\correspondence{Zixin Su at \email{suzixin.0803@bytedance.com} and \email{suzx26@tsinghua.edu.cn}}

\checkdata[Dataset]{\url{https://huggingface.co/datasets/Susan0803/visual-perception-benchmark-anon}}

\begin{document}
\maketitle

\input{sections/01_intro}
\input{sections/02_related_work}
\input{sections/03_human-level_perception_bench}
\input{sections/04_experiments}
\input{sections/06_conclusion}

\clearpage

\bibliographystyle{plainnat}
\bibliography{main}

\clearpage

\beginappendix

\input{sections/appendix/a_taxonomy}

\input{sections/appendix/b_protocol}
\input{sections/appendix/c_results}

\end{document}

%% file: sections/01_intro.tex
\section{Introduction}

Multimodal large language models (MLLMs) have demonstrated strong performance on tasks requiring substantial knowledge and multi-step reasoning, including chart interpretation and college-level problems \citep{yue2024mmmu,lu2024mathvista,wang2024measuring,zhang2024mathverse,li2023seed,yu2023mm,liu2023mmbench,fu2023mme,li2024vlmevalkit}. However, these achievements do not establish mastery of basic visual skills: MLLMs still struggle with tasks ordinary adults can easily perform, such as counting objects, following lines through intersections, reading noisy labels, and distinguishing mirrored patterns \citep{fu2024blink,tong2024eyes,kamoi2024visonlyqa,chen2026babyvision,liu2023ocrbench,masry2022chartqa,johnson2017clevr,acharya2019tallyqa}. For example, Figure~\ref{fig:overview}(a) asks how many eggs three trays can hold, requiring models to infer total capacity from the visible tray structure. Such visual judgments are necessary for reliable operation in human-facing environments, where users expect systems to interpret familiar objects and arrangements correctly. Evaluation should therefore assess not only relative performance across models but also whether they possess the basic perceptual abilities ordinary people rely on in everyday tasks. We focus on visual tasks that humans can easily perform, examining whether models reliably recover the information humans obtain from the same input. This evaluation tests a prerequisite for reliable visual assistance, rather than establishing the reliability of the entire system.

\input{sections/figures/overview_figure}

The problem is also reflected in benchmark results. Some general-purpose multimodal benchmarks such as MMMU and MathVista focus on end-to-end problem-solving ability. Their scores combine visual perception, language understanding, prior knowledge, and reasoning; some items can also be partially solved from language priors or question artifacts \citep{chen2024we,li2023pope,liu2023hallusionbench}. Perception-oriented benchmarks address the visual bottleneck more directly. For example, BLINK benchmarks traditional computer vision tasks in the setting of MLLM \citep{fu2024blink}. MMVP evaluates vision-language representations and identifies visual patterns that challenge current models \citep{tong2024eyes}. CV-Bench and VisOnlyQA focus on spatial, geometric and pure vision skills \citep{tong2024cambrian,kamoi2024visonlyqa}. BabyVision benchmarks vision skills for pre-linguistic vision \citep{chen2026babyvision}. PerceptionBench builds atomic questions for visual perception skills based on the attribution to model failure \citep{lin2026perceptionbench}. Benchmarks for scene text, documents, charts, counting, spatial reasoning, and referring expressions further show that visual perception comprises multiple distinct sub-skills \citep{singh2019textvqa,mathew2021docvqa,masry2022chartqa,acharya2019tallyqa,liu2022vsr,yu2016refcoco}. HumanSightBench is built upon the progress in atomic decomposition and human validation. Our benchmark organizes human-validated tasks into 21 fine-grained ability groups to reveal which everyday perceptual requirements remain unmet despite improvements in aggregate scores.

A benchmark for human-level visual perception should establish human solvability independently of model performance. Each item must have a well-specified target and a unique answer supported by the image alone, pass independent human verification, and resist text-only shortcuts. These criteria help distinguish gaps on human-solvable visual tasks from difficulties with specialized knowledge or complex reasoning. Model failures then reveal capability gaps on these validated tasks, without defining human-level perception or establishing the causes of failure.

We use these principles to create HumanSightBench (Figure \ref{fig:overview}). For a candidate image-question-answer tuple to be included in HumanSightBench, it must pass through five inclusion gates G1 (question clarity), G2 (answer uniqueness), G3 (visual grounding), G4 (human solvability), and G5 (text-only shortcut filtering) (Figure \ref{fig:overview}(b)). Only after candidates pass these gates do we use MLLMs to filter saturated items for diagnostic value; model performance does not define human solvability or independently justify inclusion. We collected 2{,}876 short-answer questions across seven perceptual categories and 21 subcategories (from OCR to structured visual perception to visual tracking, Figure \ref{fig:overview}(c)). On our benchmark, the best among eight representative MLLMs achieves only 65.6\% pass@3, and the best MLLM on the Counting ability solves only 27.5\% of Counting questions. This ability breakdown can guide hypotheses for targeted training-data collection, with the benchmark reserved for held-out evaluation of the relevant skills. A low score corresponds to a capability gap; to assess whether further data can close this gap, we need to run a training experiment.

Our contributions are threefold:
\begin{itemize}
  \item We propose a protocol for constructing human-easy visual perception tasks. The tasks are unambiguous, visually grounded, reliably solvable by ordinary humans, and resistant to text-only or knowledge-based shortcuts.

    \item We develop HumanSightBench, a benchmark covering seven categories of human-easy visual perception tasks. We filter ambiguity, verify answers, and check visual evidence so that errors better reflect perceptual limitations rather than task misunderstanding.

  \item We evaluate representative MLLMs on HumanSightBench and analyze performance across perceptual abilities. Despite strong results on existing multimodal benchmarks, current MLLMs still exhibit consistent weaknesses on human-easy visual perception tasks.
\end{itemize}

%% file: sections/figures/overview_figure.tex
\begin{figure}[t]
    \centering
    \includegraphics[width=\linewidth]{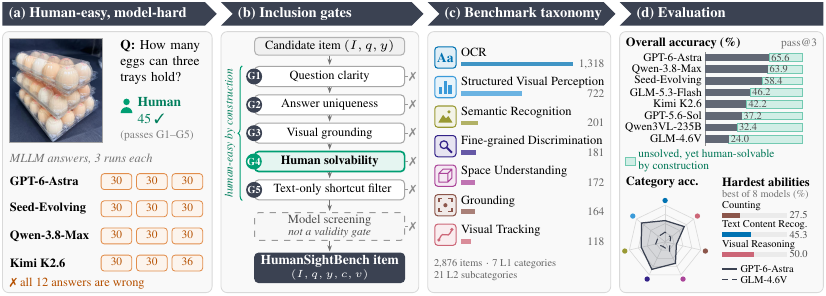}
\caption{\textbf{HumanSightBench is human-easy by construction and model-hard by evaluation.}
Items pass five validity gates before screening.
The 2{,}876-item benchmark spans 7 categories and 21 ability groups.
Among eight MLLMs, the best pass@3 is 65.6\% overall and 27.5\% on Counting.}
    \label{fig:overview}
\end{figure}

%% file: sections/02_related_work.tex
\section{Related Work}

\subsection{Multimodal Evaluation and Perceptual Entanglement}

As the general-purpose multimodal benchmark has provided a lot of information about the ability of MLLMs, the image-conditioned answer prediction is adopted as the default evaluation method for both visual question answering and diagram benchmarks at early and mid scales \citep{antol2015vqa,hudson2019gqa,kembhavi2016ai2d,lu2022scienceqa}. MMMU evaluates college-level multimodal understanding across many academic disciplines \citep{yue2024mmmu}; MathVista, MATH-Vision, and MathVerse test mathematical reasoning over visual inputs \citep{lu2024mathvista,wang2024measuring,zhang2024mathverse}. LLaVA-Bench, MMBench, MME, SEED-Bench, MM-Vet, and VLMEvalKit, etc. were proposed to evaluate the integrated ability in different application fields and model families for visual language \citep{li2023llavabench,liu2023mmbench,fu2023mme,li2023seed,yu2023mm,li2024vlmevalkit}. These benchmarks are crucial as the applications of the multimodal models require a combination of many multimodal abilities, including knowledge, language comprehension and reasoning, visual comprehension and reasoning, which brings many difficulties to analyze the root cause of the error in visual perception, including wrong visual fact perception, language ambiguity, missing background knowledge, and downstream reasoning error. A correct answer may also exploit language priors or question artifacts without using the image \citep{chen2024we,li2023pope,liu2023hallusionbench}. HumanSightBench is introduced to complement the existing end-to-end multimodal ability benchmarks and reduce confounds from external knowledge and complex reasoning when evaluating the visual perception tasks that are human-easy by construction and model-hard by evaluation.

\subsection{Visual Perception Evaluation for MLLMs}

Some works have started to investigate the visual perception bottleneck in MLLMs. Fine-grained visual perception benchmarks cover scene text, documents, charts, and OCR \citep{singh2019textvqa,biten2019stvqa,mathew2021docvqa,mathew2022infographicvqa,kafle2018dvqa,methani2020plotqa,masry2022chartqa,liu2023ocrbench,liu2025ocrbenchv2}; compositional benchmarks examine counting, spatial relations, and grounding \citep{johnson2017clevr,acharya2019tallyqa,liu2022vsr,kazemzadeh2014referit,yu2016refcoco}. BLINK reformulates classic computer vision tasks into multimodal questions and shows a large human-model gap \citep{fu2024blink}. MMVP finds the visual patterns where humans and models perform differently and which are difficult for widely used vision-language representations \citep{tong2024eyes}. CV-Bench (2D 3D) evaluates visual perception tasks in Cambrian benchmark \citep{tong2024cambrian}. VisOnlyQA evaluates the geometric and spatial perception of diagrams without external knowledge \citep{kamoi2024visonlyqa}. Other works evaluate MLLMs in visual search, visual zooming, small details, evidence dependency \citep{wu2024vstar,shen2024zoomeye,zhang2025mllmsknow,zhang2025chainoffocus,wei2026zooming,chen2026visualneedle,zhu2026visualflip,neo2025interpreting,zheng2025deepeyes}. Recently, some works define and study the meaning of "basic visual ability". BabyVision defines specific tasks to investigate early-stage human visual perception ability, including fine-grained discrimination, visual tracking, spatial perception, and visual pattern recognition \citep{chen2026babyvision}. PerceptionBench analyzes atomic perception ability according to the failure modes of MLLMs in multimodal benchmarks and defines corresponding diagnostic questions \citep{lin2026perceptionbench}. PerceptionBench also verifies visual grounding and uses independent human review. HumanSightBench builds upon these works and creates a new benchmark containing 2,876 items categorized into 21 ability groups. It empirically investigates remaining bottlenecks in eight current MLLMs and separately evaluates visual operations included in broader ability categories.

%% file: sections/03_human-level_perception_bench.tex
\section{HumanSightBench}
\label{sec:benchmark}

HumanSightBench is built around a simple benchmark principle: every item should be solvable by ordinary humans from visual evidence in the image, yet remain challenging for current MLLMs. Figure~\ref{fig:overview}(b) summarizes the inclusion gates, and Figure~\ref{fig:benchmark-construction-pipeline} details how they are carried out during collection and annotation. Our first step is to define the perceptual space to be measured. Second, we collect candidate image-question pairs. Third, only pairs that pass all five inclusion gates (G1--G5: question clarity, answer uniqueness, visual grounding, human solvability, and text-only shortcut filtering) are kept.

\subsection{Human-Level Item Definition}

Human-level solvability is a requirement for each item, rather than an aggregate baseline. Our inclusion protocol specifies an $(I, q, y, c, v)$ tuple, where $I$ is an image, $q$ is a short-answer question, $y$ is a unique verified answer, $c$ is a taxonomy path, and $v$ is a verification record that logs the set of checks that are verified. The protocol admits a candidate only if the question is well-defined, the answer is unique, the answer is supported by visual evidence in the image, three annotators independently solve it with unanimous correctness under the stated protocol, and the answer is not recoverable from text alone. Finally, model performance is used to remove saturated items, leaving items that are still challenging for the state-of-the-art MLLMs. Model failure on HumanSightBench corresponds to giving the wrong answer on a human-validated visual task under the evaluated input interface.

\subsection{Taxonomy and Coverage}
\label{sec:taxonomy}

The complete dataset is organized into 7 top-level categories (L1) and
21 subcategories (L2): Grounding, Semantic Recognition, Fine-grained
Discrimination, Visual Tracking, Space Understanding, OCR, and Structured
Visual Perception. The L2 subcategories specify abilities within each category,
such as Counting, Line Tracing, Character Recognition, and Chart Recognition.
Appendix~\ref{app:taxonomy} lists the L1 counts and full L2 names.
The taxonomy determines what we attempt to
cover; the inclusion gates determine what enters the benchmark. Some leaves are
discovery-driven because they originate from observed model failures, but model
scores alone never include an item. This separation keeps the taxonomy useful for
coverage while ensuring that the benchmark is defined by validity and
human-level solvability rather than by any single generation of models.

\subsection{Human-Level Inclusion Gates}
\label{sec:inclusion-gates}

\begin{figure}[t]
  \centering
  \includegraphics[width=\linewidth]{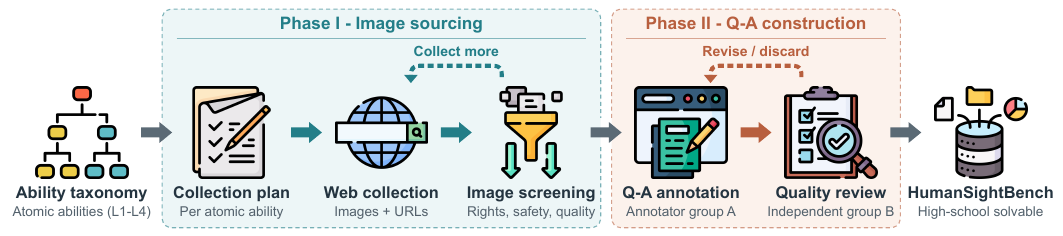}
  \caption{\textbf{Data construction pipeline of HumanSightBench.}
  The perceptual taxonomy guides collection planning, web collection, and image
  screening. One annotator group constructs question--answer pairs, and a second
  group independently reviews their quality. The feedback loops support further
  collection and the revision or removal of unsuitable pairs. The collection
  taxonomy is shown at L1--L4 granularity; evaluation uses the 7 L1 categories
  and 21 L2 groups listed in Appendix~\ref{app:taxonomy}.}
  \label{fig:benchmark-construction-pipeline}
\end{figure}

Figure~\ref{fig:benchmark-construction-pipeline} summarizes the collection and
annotation workflow.
We first collect candidate images and question ideas according to the taxonomy, drop images with copyright, privacy, low quality, unsafe, and leaking answer issues, discard candidates whose solution depends on specialized background knowledge rather than visual inspection, and finally convert the candidate images to short answer visual questions and apply the five inclusion gates below.

\textbf{G1: Question clarity.} In pilot annotation, many apparent perception failures were caused by underspecified questions. We defined a list of ambiguity type (i.e., "completeness", "occlusion", "rules regarding image boundary", "adjacency", "counted units", "ordering criteria", "output format", etc.) for each of them. These questions were iteratively edited until consistent annotation on questions specification and output format was provided by independent annotators.

\textbf{G2--G3: Answer uniqueness and visual grounding.} Each item must have a single verified answer justified by the image alone. Reviewers reject candidates requiring unsupported assumptions, world knowledge, or information outside the image. Verification records identify the elements or regions determining counting and grounding answers, and the attributes or regions supporting comparison and recognition answers. These records support answer verification and reduce reliance on hidden task interpretations.

\textbf{G4: Human solvability.} Human-level solvability is a precondition for inclusion. Ordinary human annotators answer each candidate under the same question text used for model evaluation; items that humans find inaccurate, inconsistent, or overly time-consuming are revised or removed. More details regarding the annotation protocol, including the number of annotators, inclusion criteria, time limit, the rule of image viewing and dispute resolution rule, are provided in our Appendix~\ref{app:protocol}.

\textbf{G5: Text-only shortcut filtering.} We have also filtered the text-based solvable candidates, which are ones that can be solved without image viewing \citep{chen2024we}.

\textbf{Model screening (not a validity gate).} We have also removed items saturated by MLLMs after validation. This last step does not define human-level validity; it only keeps the benchmark challenging for the models being evaluated.

\subsection{Dataset Statistics}
\label{sec:dataset-statistics}

\noindent\begin{minipage}[t]{0.32\textwidth}
  \vspace{0pt}
  \centering
  \includegraphics[width=\linewidth]{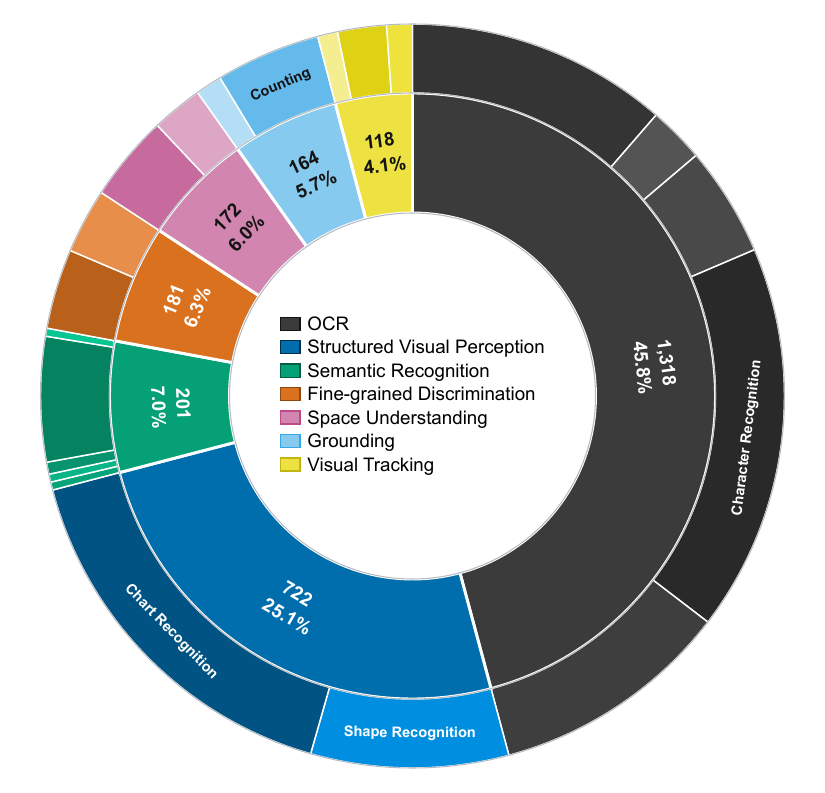}
  \captionsetup{font=footnotesize,hypcap=false}
  \captionof{figure}{\textbf{Category distribution.}} 
  \label{fig:l1-distribution}
\end{minipage}\hfill
\begin{minipage}[t]{0.64\textwidth}
\vspace{0pt}

HumanSightBench contains 2{,}876 questions across 7 L1 categories and 21 L2
subcategories. Figure~\ref{fig:l1-distribution} shows the L1 distribution, and
Appendix~\ref{app:taxonomy} provides the L1 counts and full L2 names.
OCR and Structured Visual Perception together
account for 70.9\% of the benchmark, and the largest category is 11.17 times the
size of the smallest. We do not subsample categories to force balance, because doing so
would discard valid human-level items. Instead, experiments report overall pass@3 across all items and
category-level pass@3, and
category-level conclusions are made only where the category has sufficient
coverage. This design lets HumanSightBench serve both as a broad benchmark of
human-level visual perception and as a diagnostic resource for the perceptual
conditions under which current MLLMs still fail.
\end{minipage}
\par

%% file: sections/04_experiments.tex
\section{Experiments}

\subsection{Evaluation Setup}

\paragraph{Models.}
We evaluate eight representative MLLMs: five proprietary models
(GPT-5.6-Sol, GPT-6-Astra, Qwen-3.8-Max, GLM-5.3-Flash, and Seed-Evolving)
and three open-source models
(GLM-4.6V, Kimi K2.6, and Qwen3VL-235B-Thinking).
All models are evaluated on the same 2,876 HumanSightBench questions with
consistent question text and answer-format instructions. Each instance contains
one image and one short-answer question.

\paragraph{Implementation Details.}
We use two image input settings, \textit{Resize} and \textit{No Resize}, as
reported in Table~\ref{tab:main_results}. \textit{Resize} resizes the image
before inference, while \textit{No Resize} preserves the original resolution.
The image content and question text remain unchanged. Because resizing may
affect fine-grained visual evidence, results under different settings should be
interpreted as comparisons between model--input configurations rather than as
a controlled resizing ablation. For each model--question pair, we run three
independent trials with the same image and prompt and retain the first three
responses. The prompt requests a concise answer and specifies the expected
output format when needed.

\paragraph{Evaluation Protocol and Metric.}
HumanSightBench uses open-ended short answers, so we use DeepSeek-V3.2 in
thinking mode as the answer judge instead of exact string matching. The judge
compares each final answer with the verified reference and assigns a binary
correctness label. Explanations and reasoning traces are ignored, while
semantically equivalent surface forms, including differences in capitalization,
punctuation, whitespace, and numeric formatting, are treated as correct. We
report pass@3, where a question is considered solved if at least one of the
first three trials is judged correct. Overall scores are computed over all
2,876 items, while category-level and ability-level scores are computed over
the corresponding L1 and L2 subsets. The formal definition of pass@3 is
provided in Appendix~\ref{app:evaluation-protocol}.

\input{tables/maintable}

\subsection{Overall Performance}

\paragraph{Current MLLMs retain a substantial gap on human-easy visual tasks.}
Table~\ref{tab:main_results} reports pass@3 at the overall, category, and
ability levels. Overall performance ranges from 24.0\% to 65.6\% across the
eight evaluated MLLMs. GPT-6-Astra achieves the highest overall score of
65.6\%, followed by Qwen-3.8-Max at 63.9\% and Seed-Evolving at 58.4\%.
The three open-source models obtain overall scores between 24.0\% and 42.2\%.
Even at 65.6\% pass@3, more than one third of the benchmark remains unsolved
despite every retained item having passed the human-solvability, answer-
uniqueness, visual-grounding, and text-only shortcut checks. These results
indicate that strong performance on current multimodal benchmarks does not
translate into uniformly reliable performance on visually grounded tasks that
are easy for humans under our inclusion protocol.

\paragraph{Grounding remains a consistent category-level bottleneck.}
The difficulty is highly uneven across perceptual categories. Grounding is the
lowest-scoring L1 category for all eight evaluated models, with category-level
pass@3 ranging from 11.6\% to 31.7\%. Even the best Grounding result,
31.7\% from GPT-6-Astra, remains far below the strongest results observed in
other categories. In contrast, GPT-6-Astra reaches 83.4\% on Fine-grained
Discrimination and 82.5\% on Structured Visual Perception, while
Qwen-3.8-Max reaches 70.1\% on Semantic Recognition and 62.7\% on OCR.
The large separation between Grounding and these stronger categories suggests
that the remaining perceptual difficulty is concentrated rather than uniform
across the benchmark taxonomy.

\paragraph{Models exhibit distinct perceptual profiles.}
Higher overall performance does not imply uniformly higher performance across
all categories. GPT-6-Astra achieves the best overall score and leads several
L1 categories, but Qwen-3.8-Max performs better on Semantic Recognition,
with 70.1\% versus 66.7\%, and on OCR, with 62.7\% versus 58.0\%.
Within individual models, the variation is also substantial. For example,
Qwen-3.8-Max ranges from 31.1\% on Grounding to 73.1\% on Structured Visual
Perception, while GPT-6-Astra ranges from 31.7\% on Grounding to 83.4\% on
Fine-grained Discrimination. Thus, a single aggregate score compresses
substantially different perceptual profiles and can obscure which capabilities
account for a model's strengths and weaknesses.

\paragraph{Aggregate scores motivate finer-grained diagnosis.}
L1 categories themselves still combine perceptually different operations.
OCR, for example, includes text-region localization, layout and structure
understanding, text-content recognition, character recognition, and recognition
of less-common languages. Their difficulty differs substantially within the
same category. Similarly, Visual Tracking combines line tracing, contour
recognition, and visual reasoning, which need not improve together. Category
averages can therefore conceal bottlenecks that are visible only at a finer
granularity. We next analyze the 21 L2 ability groups to localize these
remaining capability gaps.

\subsection{Ability-level Diagnosis}

\label{sec:diagnosis}

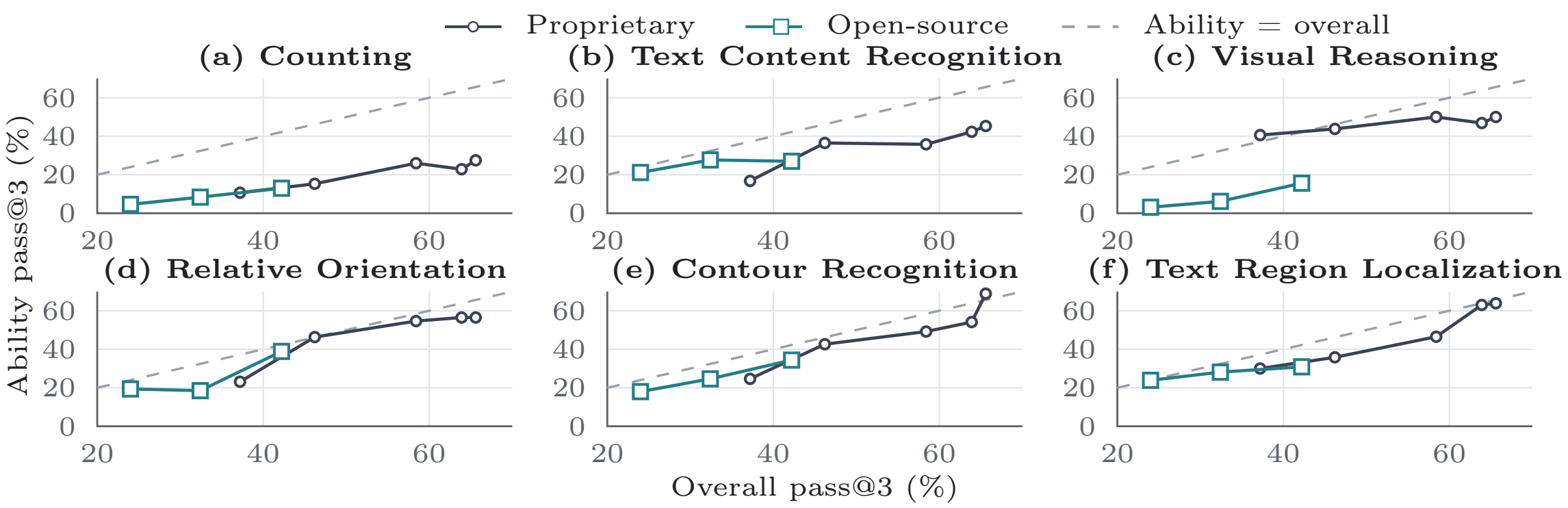
\begin{figure}[!t]
\centering
\resizebox{\linewidth}{!}{\input{figures/ability_difficulty_landscape}}
\caption{\textbf{Overall performance and the six most challenging perceptual abilities.}
Each panel plots ability-level pass@3 against overall pass@3. Lines connect
proprietary and open-source models by overall performance; the dashed line
indicates matching ability-level and overall performance.}
\label{fig:ability_landscape}
\end{figure}

\paragraph{Model failures concentrate on specific perceptual abilities.}
To localize the remaining bottlenecks, we analyze the 21 ability groups in
Table~\ref{tab:main_results}. The six lowest-scoring abilities are Counting
with a mean pass@3 of 16.0\%, Text Content Recognition at 31.6\%, Visual
Reasoning at 32.0\%, Relative Orientation and Reference Frame Transformation
at 39.2\%, Contour Recognition at 39.5\%, and Text Region Localization and
Understanding at 40.3\%. These abilities span Grounding, OCR, Visual Tracking,
and Space Understanding, indicating that the remaining difficulties are
distributed across multiple perceptual categories rather than concentrated in
a single one. Category-level aggregates can therefore mask substantial
differences among the underlying abilities.

\paragraph{Aggregate gains do not uniformly transfer to bottleneck abilities.}
Figure~\ref{fig:ability_landscape} relates performance on these six abilities
to each model's overall score. Most points remain below the equality line.
Counting shows the clearest ceiling: ability-level accuracy ranges from 4.6\%
to 27.5\% even as overall accuracy rises from 24.0\% to 65.6\%.
Text Content Recognition and Visual Reasoning also remain difficult, with best
scores of 45.3\% and 50.0\%. The trajectories are non-monotonic:
Qwen-3.8-Max remains below Seed-Evolving on Counting and Visual Reasoning, while
GPT-6-Astra trails Qwen-3.8-Max on Abstract Concept Grounding and Color
Perception. Stronger overall performance does not imply uniformly stronger
atomic perception.

\paragraph{Shared failures reveal common bottlenecks.}
Before the GPT-6-Astra update, 461 of 2,876 items (16.0\%) were unsolved by all
eight models across all three runs. The within-ability all-model failure rate is
61.8\% for Counting, 37.2\% for Text Content Recognition, 28.1\% for Visual
Reasoning, and 24.6\% for Contour Recognition. These rates show that the weakest
abilities are not driven only by a few poorly performing models, but include
items that remain difficult across diverse model families and architectures.
Counting is particularly notable, with more than half of its items unsolved by
every model in this diagnostic snapshot. Such shared failures provide stronger
evidence of common capability limitations than errors concentrated in a single
model or model family. These rates describe the earlier snapshot, not the
updated model set in Table~\ref{tab:main_results}.

\begin{figure}[!t]
\centering
\includegraphics[width=0.90\linewidth]{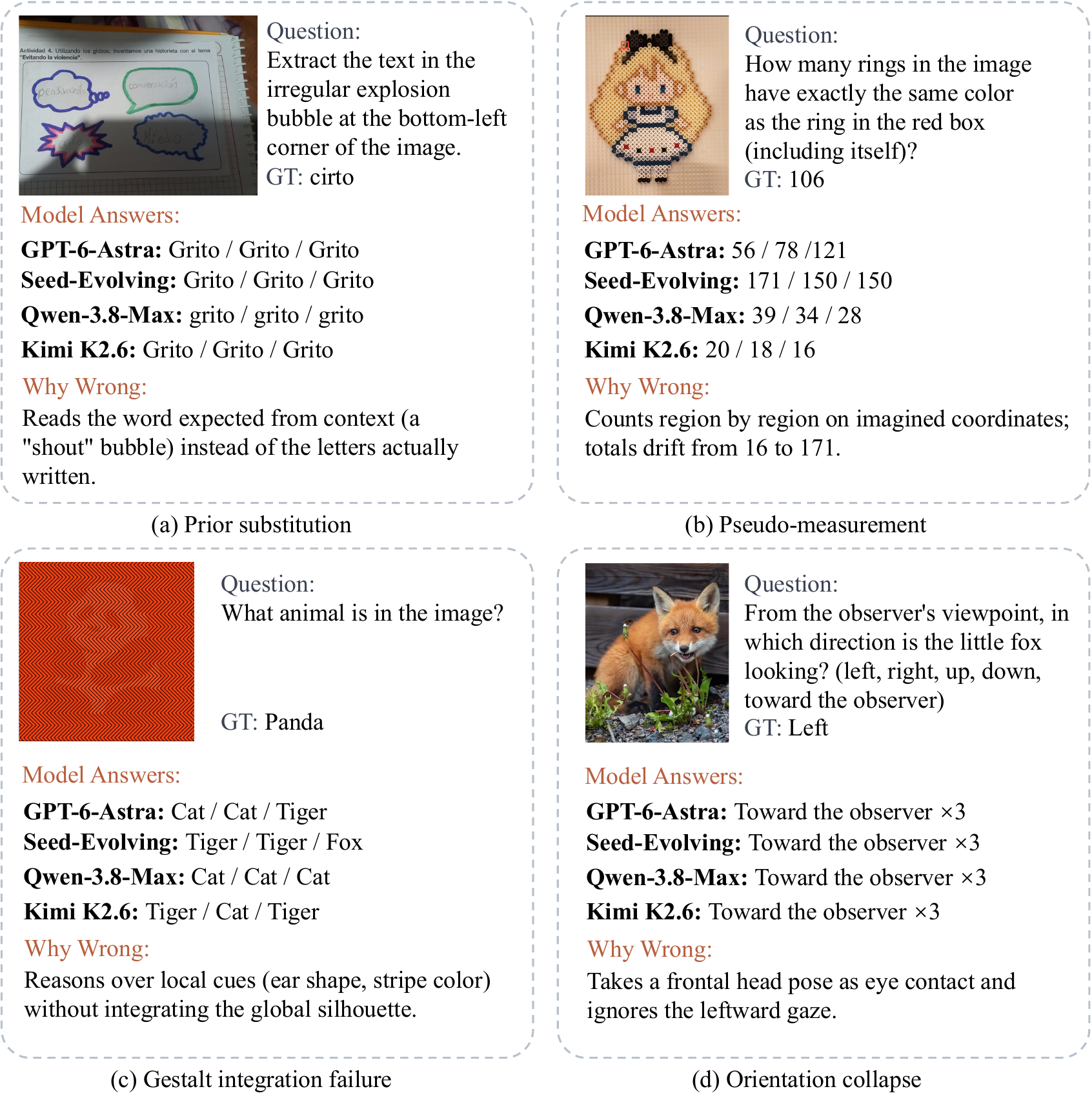}
\caption{\textbf{Four illustrative failure patterns on HumanSightBench.} (a) prior substitution; (b) pseudo-measurement; (c) gestalt integration failure; (d) orientation collapse. These cases come from the earlier diagnostic snapshot. Each displayed example is failed by Qwen-3.8-Max, Seed-Evolving, GPT-6-Astra, and Kimi K2.6. The descriptions summarize these outputs without establishing causal mechanisms or their prevalence.}
\label{fig:failure_cases}
\end{figure}

\paragraph{Models exhibit distinct ability-level profiles.}
GPT-6-Astra ranks first outright in 11 of the 21 abilities and ties for first in
3. Qwen-3.8-Max ranks first outright in 5 abilities and ties for first in 2,
while Seed-Evolving ranks first outright in 2 abilities and ties for first in
1. The distribution of ability-level leaders shows that no model dominates
uniformly across the taxonomy or across different perceptual operations.
Thus, strong overall performance can arise from different combinations of
perceptual strengths rather than uniform gains across abilities, and models
with similar aggregate scores may still differ substantially in where their
strengths and weaknesses lie. Appendix~\ref{app:additional-results} reports
group sizes and alternative weighting schemes to support interpretation of
these comparisons.

\subsection{Failure Mode Analysis}
\label{sec:failure}

Based on the previous diagnostic snapshot, we study the traces of the three leading proprietary models: Qwen-3.8-Max, Seed-Evolving, and GPT-6-Astra, and the leading open-source model: Kimi K2.6, on shared failures. Figure~\ref{fig:failure_cases} shows 4 cases. (a) \textbf{Prior substitution}: the model replaces the literal visual content with the expected form, e.g., after detecting a misspelled word, it gives the corrected spelling instead of the literal text. (b) \textbf{Pseudo-measurement}: in the ring-counting task, the displayed traces contain coordinate-based counting descriptions with incorrect totals that vary across runs and models. (c) \textbf{Gestalt integration failure}: local information is extracted without integrating global information, e.g., an animal hidden in zigzag stripes is ignored. (d) \textbf{Orientation collapse}: continuous orientations are collapsed into canonical labels, e.g., the model mistakes a frontal head with averted gaze for looking toward the observer.

These examples inspire us to explore how we could have an agent return to the visual evidence during its reasoning process instead of solely referring to the text description. These examples do not tell us how frequent this pattern is, or establish that additional reasoning causes errors.

\begin{figure}[t]
\centering
\includegraphics[width=\linewidth]{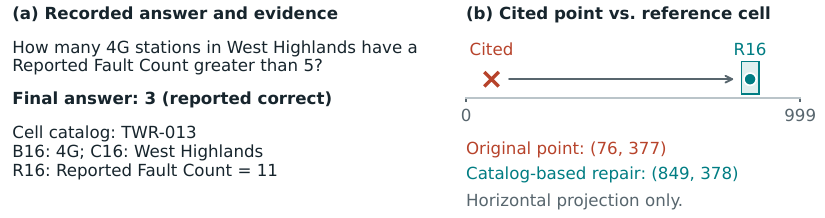}
\caption{\textbf{Correct answer, misplaced explicit evidence.} An external table-QA example where the reported answer is correct despite an incorrect evidence location.}
\label{fig:answer-evidence-case}
\end{figure}

\paragraph{Answer correctness does not certify evidence localization.}
Figure~\ref{fig:answer-evidence-case} illustrates a boundary of answer-only scoring: a reported correct answer coexists with a misplaced explicit evidence pointer. The repair is post-hoc, not a new model prediction. Without the complete original table, we cannot independently verify the reported answer. This external illustration motivates separating answer accuracy from explicit evidence localization; it establishes neither the frequency of such errors on HumanSightBench nor the model's internal evidence use.

\paragraph{Evaluation boundaries.}
Our inclusion gates reduce confounds but do not isolate perception as the causal source of incorrect answers. Incorrectness can also stem from task interpretation, reasoning, output conventions, or judge error. Likewise, the accuracy of localization may require consideration of coordinate conventions and target regions before it can be attributed to perception. The pass@3 metric measures at-least-once success across three trials, not repeatable correctness. Human solvability is confirmed following our annotation protocol, not estimated for the general population. Model-based screening and unequal category coverage make the scores diagnostic of retained items rather than estimates of everyday error prevalence. Models were evaluated under two preprocessing options labeled ``Resize'' and ``No Resize'' in Table \ref{tab:main_results}. These comparisons combine model and input-interface differences. Human annotators were allowed to zoom the original images, therefore these results are not a controlled comparison at matched viewing conditions.

%% file: tables/maintable.tex
\definecolor{myyellow}{RGB}{253,242,208}

\begin{table*}[!t]
\renewcommand\arraystretch{1.12}
\setlength{\tabcolsep}{2pt}
\setlength{\aboverulesep}{1.3pt}
\setlength{\belowrulesep}{1.3pt}
\centering
\caption{\textbf{Performance of MLLMs on HumanSightBench.} We evaluate five proprietary MLLMs and three open-source MLLMs. The best results for each question type are marked in \textbf{bold}. Reported values represent pass@3 accuracy, i.e.\ the proportion of samples solved in at least one of the first three trials. Subtype names are abbreviated; full names appear in Appendix~\ref{app:taxonomy}.}
\label{tab:main_results}
\begingroup
\fontsize{8}{9.4}\selectfont
\begin{tabular*}{\linewidth}{@{\extracolsep{\fill}} c l *{8}{c} @{}}
\toprule
\multirow{3}{*}{\textbf{Type}} &
\multirow{3}{*}{\textbf{Subtype}} &
\multicolumn{5}{c}{\textbf{Proprietary MLLMs}} &
\multicolumn{3}{c}{\textbf{Open-source MLLMs}} \\
\cmidrule(lr){3-7} \cmidrule(lr){8-10}
& &
\multicolumn{3}{c}{No Resize} &
\multicolumn{2}{c}{Resize} &
\multicolumn{2}{c}{No Resize} &
Resize \\
\cmidrule(lr){3-7} \cmidrule(lr){8-10}
& &
\makecell{\bfseries GPT-5.6\\\bfseries Sol} & \makecell{\bfseries GPT-6\\\bfseries Astra} & \makecell{\bfseries GLM-5.3\\\bfseries Flash} & \makecell{\bfseries Seed\\\bfseries Evolving} & \makecell{\bfseries Qwen-3.8\\\bfseries Max} &
\makecell{\bfseries GLM\\\bfseries 4.6V} & \makecell{\bfseries Kimi\\\bfseries K2.6} & \makecell{\bfseries Qwen3VL\\\bfseries 235B\\\bfseries Thinking} \\
\midrule

\multirow{3}{*}{\makecell{Grounding}}
& Abstract concept & 15.2 & 48.5 & 36.4 & 51.5 & \textbf{63.6} & 48.5 & 36.4 & 42.4 \\
& Counting & 10.7 & \textbf{27.5} & 15.3 & 26.0 & 22.9 & 4.6 & 13.0 & 8.4 \\
\cmidrule(lr){2-10}
& \cellcolor{blue!5}\textbf{Category overall}
& \cellcolor{blue!5}11.6 & \cellcolor{blue!5}\textbf{31.7} & \cellcolor{blue!5}19.5 & \cellcolor{blue!5}31.1 & \cellcolor{blue!5}31.1
& \cellcolor{blue!5}13.4 & \cellcolor{blue!5}17.7 & \cellcolor{blue!5}15.2 \\
\midrule
\multirow{6}{*}{\makecell{Semantic\\Recognition}}
& Basic Recognition & 30.0 & \textbf{90.0} & 10.0 & 50.0 & \textbf{90.0} & 20.0 & 50.0 & 40.0 \\
& Actions \& interact. & 50.0 & 60.0 & 60.0 & 50.0 & \textbf{90.0} & 40.0 & 30.0 & 10.0 \\
& States \& attrib. & 46.7 & 60.0 & 53.3 & \textbf{66.7} & 60.0 & 33.3 & 53.3 & 53.3 \\
& Color Perception & 37.2 & 65.4 & 41.7 & 53.8 & \textbf{69.2} & 25.6 & 35.3 & 30.1 \\
& Semantic Relations & 40.0 & \textbf{80.0} & 40.0 & 60.0 & 60.0 & 20.0 & 20.0 & 30.0 \\
\cmidrule(lr){2-10}
& \cellcolor{blue!5}\textbf{Category overall}
& \cellcolor{blue!5}38.3 & \cellcolor{blue!5}66.7 & \cellcolor{blue!5}41.8 & \cellcolor{blue!5}54.7 & \cellcolor{blue!5}\textbf{70.1}
& \cellcolor{blue!5}26.4 & \cellcolor{blue!5}36.3 & \cellcolor{blue!5}31.3 \\
\midrule
\multirow{3}{*}{\makecell{Fine-grained\\discrim.}}
& Basic Recognition & 40.0 & \textbf{76.0} & 40.0 & 61.0 & 61.0 & 14.0 & 35.0 & 23.0 \\
& Grid Recognition & 54.3 & \textbf{92.6} & 42.0 & 75.3 & 63.0 & 13.6 & 29.6 & 14.8 \\
\cmidrule(lr){2-10}
& \cellcolor{blue!5}\textbf{Category overall}
& \cellcolor{blue!5}46.4 & \cellcolor{blue!5}\textbf{83.4} & \cellcolor{blue!5}40.9 & \cellcolor{blue!5}67.4 & \cellcolor{blue!5}61.9
& \cellcolor{blue!5}13.8 & \cellcolor{blue!5}32.6 & \cellcolor{blue!5}19.3 \\
\midrule
\multirow{4}{*}{\makecell{Visual\\Tracking}}
& Line Tracing & 28.0 & \textbf{84.0} & 48.0 & 68.0 & 80.0 & 44.0 & 48.0 & 32.0 \\
& Contour Recognition & 24.6 & \textbf{68.9} & 42.6 & 49.2 & 54.1 & 18.0 & 34.4 & 24.6 \\
& Visual Reasoning & 40.6 & \textbf{50.0} & 43.8 & \textbf{50.0} & 46.9 & 3.1 & 15.6 & 6.2 \\
\cmidrule(lr){2-10}
& \cellcolor{blue!5}\textbf{Category overall}
& \cellcolor{blue!5}29.7 & \cellcolor{blue!5}\textbf{66.9} & \cellcolor{blue!5}44.1 & \cellcolor{blue!5}53.4 & \cellcolor{blue!5}57.6
& \cellcolor{blue!5}19.5 & \cellcolor{blue!5}32.2 & \cellcolor{blue!5}21.2 \\
\midrule
\multirow{3}{*}{\makecell{Space\\underst.}}
& Rel. orient. / frame transf. & 23.1 & \textbf{56.5} & 46.3 & 54.6 & \textbf{56.5} & 19.4 & 38.9 & 18.5 \\
& Mirror perception & 50.0 & \textbf{79.7} & 70.3 & 70.3 & 78.1 & 42.2 & 54.7 & 39.1 \\
\cmidrule(lr){2-10}
& \cellcolor{blue!5}\textbf{Category overall}
& \cellcolor{blue!5}33.1 & \cellcolor{blue!5}\textbf{65.1} & \cellcolor{blue!5}55.2 & \cellcolor{blue!5}60.5 & \cellcolor{blue!5}64.5
& \cellcolor{blue!5}27.9 & \cellcolor{blue!5}44.8 & \cellcolor{blue!5}26.2 \\
\midrule
\multirow{6}{*}{\makecell{OCR}}
& Text region loc. / und. & 30.0 & \textbf{63.9} & 35.8 & 46.5 & 63.0 & 23.9 & 30.9 & 28.1 \\
& Text layout \& structure & 50.0 & 81.4 & 61.4 & 70.0 & \textbf{84.3} & 31.4 & 61.4 & 48.6 \\
& Text content recog. & 16.8 & \textbf{45.3} & 36.5 & 35.8 & 42.3 & 21.2 & 27.0 & 27.7 \\
& Character recog. & 19.6 & 53.9 & 40.3 & 59.5 & \textbf{66.9} & 25.8 & 58.7 & 47.1 \\
& Less-common-lang. OCR & 43.0 & 58.3 & 53.0 & \textbf{60.7} & 60.0 & 32.7 & 51.3 & 41.3 \\
\cmidrule(lr){2-10}
& \cellcolor{blue!5}\textbf{Category overall}
& \cellcolor{blue!5}28.8 & \cellcolor{blue!5}58.0 & \cellcolor{blue!5}42.8 & \cellcolor{blue!5}54.6 & \cellcolor{blue!5}\textbf{62.7}
& \cellcolor{blue!5}26.7 & \cellcolor{blue!5}47.0 & \cellcolor{blue!5}39.2 \\
\midrule
\multirow{3}{*}{\makecell{Structured\\visual percept.}}
& Shape Recognition & 61.9 & \textbf{89.9} & 65.2 & 79.8 & 84.2 & 27.9 & 53.4 & 38.1 \\
& Chart Recognition & 55.8 & \textbf{78.7} & 56.4 & 66.1 & 67.4 & 20.6 & 39.4 & 27.2 \\
\cmidrule(lr){2-10}
& \cellcolor{blue!5}\textbf{Category overall}
& \cellcolor{blue!5}57.9 & \cellcolor{blue!5}\textbf{82.5} & \cellcolor{blue!5}59.4 & \cellcolor{blue!5}70.8 & \cellcolor{blue!5}73.1
& \cellcolor{blue!5}23.1 & \cellcolor{blue!5}44.2 & \cellcolor{blue!5}30.9 \\
\midrule
\multirow{1}{*}{\makecell{All}}
& \cellcolor{myyellow}\textbf{Overall}
& \cellcolor{myyellow}37.2 & \cellcolor{myyellow}\textbf{65.6} & \cellcolor{myyellow}46.2 & \cellcolor{myyellow}58.4 & \cellcolor{myyellow}63.9
& \cellcolor{myyellow}24.0 & \cellcolor{myyellow}42.2 & \cellcolor{myyellow}32.4 \\
\bottomrule
\end{tabular*}
\endgroup
\end{table*}

%% file: figures/ability_difficulty_landscape.tex
\definecolor{abilityGrid}{HTML}{E2E5E9}
\definecolor{abilityAxis}{HTML}{5F6368}
\definecolor{abilityText}{HTML}{202124}
\definecolor{abilityRef}{HTML}{9AA0A6}
\definecolor{abilityProp}{HTML}{3B4252}
\definecolor{abilityOpen}{HTML}{237E88}
\begin{tikzpicture}[font=\fontsize{6.5}{7}\selectfont]
\path[use as bounding box] (0,0) rectangle (13.45,4.05);
\draw[abilityProp,line width=0.65pt] (3.650,3.94) -- (4.120,3.94);
\filldraw[fill=white,draw=abilityProp] (3.885,3.94) circle (1.15pt);
\node[anchor=west,text=abilityText] at (4.200,3.94) {Proprietary};
\draw[abilityOpen,line width=0.65pt] (6.150,3.94) -- (6.620,3.94);
\filldraw[fill=white,draw=abilityOpen] (6.385,3.88) rectangle (6.505,4.00);
\node[anchor=west,text=abilityText] at (6.700,3.94) {Open-source};
\draw[abilityRef,dashed,line width=0.65pt] (8.780,3.94) -- (9.250,3.94);
\node[anchor=west,text=abilityText] at (9.330,3.94) {Ability = overall};
\node[text=abilityText] at (6.725,0.10) {Overall pass@3 (\%)};
\node[text=abilityText,font=\fontsize{6.5}{7}\selectfont\bfseries] at (2.485,3.680) {(a) Counting};
\draw[abilityGrid,line width=0.35pt] (0.760,2.390) -- (0.760,3.510);
\node[anchor=north,text=abilityAxis] at (0.760,2.360) {20};
\draw[abilityGrid,line width=0.35pt] (2.140,2.390) -- (2.140,3.510);
\node[anchor=north,text=abilityAxis] at (2.140,2.360) {40};
\draw[abilityGrid,line width=0.35pt] (3.520,2.390) -- (3.520,3.510);
\node[anchor=north,text=abilityAxis] at (3.520,2.360) {60};
\draw[abilityGrid,line width=0.35pt] (0.760,2.390) -- (4.210,2.390);
\node[anchor=east,text=abilityAxis] at (0.705,2.390) {0};
\draw[abilityGrid,line width=0.35pt] (0.760,2.710) -- (4.210,2.710);
\node[anchor=east,text=abilityAxis] at (0.705,2.710) {20};
\draw[abilityGrid,line width=0.35pt] (0.760,3.030) -- (4.210,3.030);
\node[anchor=east,text=abilityAxis] at (0.705,3.030) {40};
\draw[abilityGrid,line width=0.35pt] (0.760,3.350) -- (4.210,3.350);
\node[anchor=east,text=abilityAxis] at (0.705,3.350) {60};
\draw[abilityAxis,line width=0.45pt] (4.210,2.390) -- (0.760,2.390) -- (0.760,3.510);
\draw[abilityRef,dashed,line width=0.55pt] (0.760,2.710) -- (4.210,3.510);
\draw[abilityProp,line width=0.75pt] (1.947,2.561) -- (2.568,2.635) -- (3.410,2.806) -- (3.789,2.756) -- (3.906,2.830);
\filldraw[fill=white,draw=abilityProp,line width=0.65pt] (1.947,2.561) circle (1.15pt);
\filldraw[fill=white,draw=abilityProp,line width=0.65pt] (2.568,2.635) circle (1.15pt);
\filldraw[fill=white,draw=abilityProp,line width=0.65pt] (3.410,2.806) circle (1.15pt);
\filldraw[fill=white,draw=abilityProp,line width=0.65pt] (3.789,2.756) circle (1.15pt);
\filldraw[fill=white,draw=abilityProp,line width=0.65pt] (3.906,2.830) circle (1.15pt);
\draw[abilityOpen,line width=0.75pt] (1.036,2.464) -- (1.616,2.524) -- (2.292,2.598);
\filldraw[fill=white,draw=abilityOpen,line width=0.65pt] (0.976,2.404) rectangle (1.096,2.524);
\filldraw[fill=white,draw=abilityOpen,line width=0.65pt] (1.556,2.464) rectangle (1.676,2.584);
\filldraw[fill=white,draw=abilityOpen,line width=0.65pt] (2.232,2.538) rectangle (2.352,2.658);
\node[text=abilityText,font=\fontsize{6.5}{7}\selectfont\bfseries] at (6.725,3.680) {(b) Text Content Recognition};
\draw[abilityGrid,line width=0.35pt] (5.000,2.390) -- (5.000,3.510);
\node[anchor=north,text=abilityAxis] at (5.000,2.360) {20};
\draw[abilityGrid,line width=0.35pt] (6.380,2.390) -- (6.380,3.510);
\node[anchor=north,text=abilityAxis] at (6.380,2.360) {40};
\draw[abilityGrid,line width=0.35pt] (7.760,2.390) -- (7.760,3.510);
\node[anchor=north,text=abilityAxis] at (7.760,2.360) {60};
\draw[abilityGrid,line width=0.35pt] (5.000,2.390) -- (8.450,2.390);
\node[anchor=east,text=abilityAxis] at (4.945,2.390) {0};
\draw[abilityGrid,line width=0.35pt] (5.000,2.710) -- (8.450,2.710);
\node[anchor=east,text=abilityAxis] at (4.945,2.710) {20};
\draw[abilityGrid,line width=0.35pt] (5.000,3.030) -- (8.450,3.030);
\node[anchor=east,text=abilityAxis] at (4.945,3.030) {40};
\draw[abilityGrid,line width=0.35pt] (5.000,3.350) -- (8.450,3.350);
\node[anchor=east,text=abilityAxis] at (4.945,3.350) {60};
\draw[abilityAxis,line width=0.45pt] (8.450,2.390) -- (5.000,2.390) -- (5.000,3.510);
\draw[abilityRef,dashed,line width=0.55pt] (5.000,2.710) -- (8.450,3.510);
\draw[abilityProp,line width=0.75pt] (6.187,2.659) -- (6.808,2.974) -- (7.650,2.963) -- (8.029,3.067) -- (8.146,3.115);
\filldraw[fill=white,draw=abilityProp,line width=0.65pt] (6.187,2.659) circle (1.15pt);
\filldraw[fill=white,draw=abilityProp,line width=0.65pt] (6.808,2.974) circle (1.15pt);
\filldraw[fill=white,draw=abilityProp,line width=0.65pt] (7.650,2.963) circle (1.15pt);
\filldraw[fill=white,draw=abilityProp,line width=0.65pt] (8.029,3.067) circle (1.15pt);
\filldraw[fill=white,draw=abilityProp,line width=0.65pt] (8.146,3.115) circle (1.15pt);
\draw[abilityOpen,line width=0.75pt] (5.276,2.729) -- (5.856,2.833) -- (6.532,2.822);
\filldraw[fill=white,draw=abilityOpen,line width=0.65pt] (5.216,2.669) rectangle (5.336,2.789);
\filldraw[fill=white,draw=abilityOpen,line width=0.65pt] (5.796,2.773) rectangle (5.916,2.893);
\filldraw[fill=white,draw=abilityOpen,line width=0.65pt] (6.472,2.762) rectangle (6.592,2.882);
\node[text=abilityText,font=\fontsize{6.5}{7}\selectfont\bfseries] at (10.965,3.680) {(c) Visual Reasoning};
\draw[abilityGrid,line width=0.35pt] (9.240,2.390) -- (9.240,3.510);
\node[anchor=north,text=abilityAxis] at (9.240,2.360) {20};
\draw[abilityGrid,line width=0.35pt] (10.620,2.390) -- (10.620,3.510);
\node[anchor=north,text=abilityAxis] at (10.620,2.360) {40};
\draw[abilityGrid,line width=0.35pt] (12.000,2.390) -- (12.000,3.510);
\node[anchor=north,text=abilityAxis] at (12.000,2.360) {60};
\draw[abilityGrid,line width=0.35pt] (9.240,2.390) -- (12.690,2.390);
\node[anchor=east,text=abilityAxis] at (9.185,2.390) {0};
\draw[abilityGrid,line width=0.35pt] (9.240,2.710) -- (12.690,2.710);
\node[anchor=east,text=abilityAxis] at (9.185,2.710) {20};
\draw[abilityGrid,line width=0.35pt] (9.240,3.030) -- (12.690,3.030);
\node[anchor=east,text=abilityAxis] at (9.185,3.030) {40};
\draw[abilityGrid,line width=0.35pt] (9.240,3.350) -- (12.690,3.350);
\node[anchor=east,text=abilityAxis] at (9.185,3.350) {60};
\draw[abilityAxis,line width=0.45pt] (12.690,2.390) -- (9.240,2.390) -- (9.240,3.510);
\draw[abilityRef,dashed,line width=0.55pt] (9.240,2.710) -- (12.690,3.510);
\draw[abilityProp,line width=0.75pt] (10.427,3.040) -- (11.048,3.091) -- (11.890,3.190) -- (12.269,3.140) -- (12.386,3.190);
\filldraw[fill=white,draw=abilityProp,line width=0.65pt] (10.427,3.040) circle (1.15pt);
\filldraw[fill=white,draw=abilityProp,line width=0.65pt] (11.048,3.091) circle (1.15pt);
\filldraw[fill=white,draw=abilityProp,line width=0.65pt] (11.890,3.190) circle (1.15pt);
\filldraw[fill=white,draw=abilityProp,line width=0.65pt] (12.269,3.140) circle (1.15pt);
\filldraw[fill=white,draw=abilityProp,line width=0.65pt] (12.386,3.190) circle (1.15pt);
\draw[abilityOpen,line width=0.75pt] (9.516,2.440) -- (10.096,2.489) -- (10.772,2.640);
\filldraw[fill=white,draw=abilityOpen,line width=0.65pt] (9.456,2.380) rectangle (9.576,2.500);
\filldraw[fill=white,draw=abilityOpen,line width=0.65pt] (10.036,2.429) rectangle (10.156,2.549);
\filldraw[fill=white,draw=abilityOpen,line width=0.65pt] (10.712,2.580) rectangle (10.832,2.700);
\node[text=abilityText,font=\fontsize{6.5}{7}\selectfont\bfseries] at (2.485,1.910) {(d) Relative Orientation};
\draw[abilityGrid,line width=0.35pt] (0.760,0.620) -- (0.760,1.740);
\node[anchor=north,text=abilityAxis] at (0.760,0.590) {20};
\draw[abilityGrid,line width=0.35pt] (2.140,0.620) -- (2.140,1.740);
\node[anchor=north,text=abilityAxis] at (2.140,0.590) {40};
\draw[abilityGrid,line width=0.35pt] (3.520,0.620) -- (3.520,1.740);
\node[anchor=north,text=abilityAxis] at (3.520,0.590) {60};
\draw[abilityGrid,line width=0.35pt] (0.760,0.620) -- (4.210,0.620);
\node[anchor=east,text=abilityAxis] at (0.705,0.620) {0};
\draw[abilityGrid,line width=0.35pt] (0.760,0.940) -- (4.210,0.940);
\node[anchor=east,text=abilityAxis] at (0.705,0.940) {20};
\draw[abilityGrid,line width=0.35pt] (0.760,1.260) -- (4.210,1.260);
\node[anchor=east,text=abilityAxis] at (0.705,1.260) {40};
\draw[abilityGrid,line width=0.35pt] (0.760,1.580) -- (4.210,1.580);
\node[anchor=east,text=abilityAxis] at (0.705,1.580) {60};
\draw[abilityAxis,line width=0.45pt] (4.210,0.620) -- (0.760,0.620) -- (0.760,1.740);
\draw[abilityRef,dashed,line width=0.55pt] (0.760,0.940) -- (4.210,1.740);
\draw[abilityProp,line width=0.75pt] (1.947,0.990) -- (2.568,1.361) -- (3.410,1.494) -- (3.789,1.524) -- (3.906,1.524);
\filldraw[fill=white,draw=abilityProp,line width=0.65pt] (1.947,0.990) circle (1.15pt);
\filldraw[fill=white,draw=abilityProp,line width=0.65pt] (2.568,1.361) circle (1.15pt);
\filldraw[fill=white,draw=abilityProp,line width=0.65pt] (3.410,1.494) circle (1.15pt);
\filldraw[fill=white,draw=abilityProp,line width=0.65pt] (3.789,1.524) circle (1.15pt);
\filldraw[fill=white,draw=abilityProp,line width=0.65pt] (3.906,1.524) circle (1.15pt);
\draw[abilityOpen,line width=0.75pt] (1.036,0.930) -- (1.616,0.916) -- (2.292,1.242);
\filldraw[fill=white,draw=abilityOpen,line width=0.65pt] (0.976,0.870) rectangle (1.096,0.990);
\filldraw[fill=white,draw=abilityOpen,line width=0.65pt] (1.556,0.856) rectangle (1.676,0.976);
\filldraw[fill=white,draw=abilityOpen,line width=0.65pt] (2.232,1.182) rectangle (2.352,1.302);
\node[text=abilityText,font=\fontsize{6.5}{7}\selectfont\bfseries] at (6.725,1.910) {(e) Contour Recognition};
\draw[abilityGrid,line width=0.35pt] (5.000,0.620) -- (5.000,1.740);
\node[anchor=north,text=abilityAxis] at (5.000,0.590) {20};
\draw[abilityGrid,line width=0.35pt] (6.380,0.620) -- (6.380,1.740);
\node[anchor=north,text=abilityAxis] at (6.380,0.590) {40};
\draw[abilityGrid,line width=0.35pt] (7.760,0.620) -- (7.760,1.740);
\node[anchor=north,text=abilityAxis] at (7.760,0.590) {60};
\draw[abilityGrid,line width=0.35pt] (5.000,0.620) -- (8.450,0.620);
\node[anchor=east,text=abilityAxis] at (4.945,0.620) {0};
\draw[abilityGrid,line width=0.35pt] (5.000,0.940) -- (8.450,0.940);
\node[anchor=east,text=abilityAxis] at (4.945,0.940) {20};
\draw[abilityGrid,line width=0.35pt] (5.000,1.260) -- (8.450,1.260);
\node[anchor=east,text=abilityAxis] at (4.945,1.260) {40};
\draw[abilityGrid,line width=0.35pt] (5.000,1.580) -- (8.450,1.580);
\node[anchor=east,text=abilityAxis] at (4.945,1.580) {60};
\draw[abilityAxis,line width=0.45pt] (8.450,0.620) -- (5.000,0.620) -- (5.000,1.740);
\draw[abilityRef,dashed,line width=0.55pt] (5.000,0.940) -- (8.450,1.740);
\draw[abilityProp,line width=0.75pt] (6.187,1.014) -- (6.808,1.302) -- (7.650,1.407) -- (8.029,1.486) -- (8.146,1.722);
\filldraw[fill=white,draw=abilityProp,line width=0.65pt] (6.187,1.014) circle (1.15pt);
\filldraw[fill=white,draw=abilityProp,line width=0.65pt] (6.808,1.302) circle (1.15pt);
\filldraw[fill=white,draw=abilityProp,line width=0.65pt] (7.650,1.407) circle (1.15pt);
\filldraw[fill=white,draw=abilityProp,line width=0.65pt] (8.029,1.486) circle (1.15pt);
\filldraw[fill=white,draw=abilityProp,line width=0.65pt] (8.146,1.722) circle (1.15pt);
\draw[abilityOpen,line width=0.75pt] (5.276,0.908) -- (5.856,1.014) -- (6.532,1.170);
\filldraw[fill=white,draw=abilityOpen,line width=0.65pt] (5.216,0.848) rectangle (5.336,0.968);
\filldraw[fill=white,draw=abilityOpen,line width=0.65pt] (5.796,0.954) rectangle (5.916,1.074);
\filldraw[fill=white,draw=abilityOpen,line width=0.65pt] (6.472,1.110) rectangle (6.592,1.230);
\node[text=abilityText,font=\fontsize{6.5}{7}\selectfont\bfseries] at (10.965,1.910) {(f) Text Region Localization};
\draw[abilityGrid,line width=0.35pt] (9.240,0.620) -- (9.240,1.740);
\node[anchor=north,text=abilityAxis] at (9.240,0.590) {20};
\draw[abilityGrid,line width=0.35pt] (10.620,0.620) -- (10.620,1.740);
\node[anchor=north,text=abilityAxis] at (10.620,0.590) {40};
\draw[abilityGrid,line width=0.35pt] (12.000,0.620) -- (12.000,1.740);
\node[anchor=north,text=abilityAxis] at (12.000,0.590) {60};
\draw[abilityGrid,line width=0.35pt] (9.240,0.620) -- (12.690,0.620);
\node[anchor=east,text=abilityAxis] at (9.185,0.620) {0};
\draw[abilityGrid,line width=0.35pt] (9.240,0.940) -- (12.690,0.940);
\node[anchor=east,text=abilityAxis] at (9.185,0.940) {20};
\draw[abilityGrid,line width=0.35pt] (9.240,1.260) -- (12.690,1.260);
\node[anchor=east,text=abilityAxis] at (9.185,1.260) {40};
\draw[abilityGrid,line width=0.35pt] (9.240,1.580) -- (12.690,1.580);
\node[anchor=east,text=abilityAxis] at (9.185,1.580) {60};
\draw[abilityAxis,line width=0.45pt] (12.690,0.620) -- (9.240,0.620) -- (9.240,1.740);
\draw[abilityRef,dashed,line width=0.55pt] (9.240,0.940) -- (12.690,1.740);
\draw[abilityProp,line width=0.75pt] (10.427,1.100) -- (11.048,1.193) -- (11.890,1.364) -- (12.269,1.628) -- (12.386,1.642);
\filldraw[fill=white,draw=abilityProp,line width=0.65pt] (10.427,1.100) circle (1.15pt);
\filldraw[fill=white,draw=abilityProp,line width=0.65pt] (11.048,1.193) circle (1.15pt);
\filldraw[fill=white,draw=abilityProp,line width=0.65pt] (11.890,1.364) circle (1.15pt);
\filldraw[fill=white,draw=abilityProp,line width=0.65pt] (12.269,1.628) circle (1.15pt);
\filldraw[fill=white,draw=abilityProp,line width=0.65pt] (12.386,1.642) circle (1.15pt);
\draw[abilityOpen,line width=0.75pt] (9.516,1.002) -- (10.096,1.070) -- (10.772,1.114);
\filldraw[fill=white,draw=abilityOpen,line width=0.65pt] (9.456,0.942) rectangle (9.576,1.062);
\filldraw[fill=white,draw=abilityOpen,line width=0.65pt] (10.036,1.010) rectangle (10.156,1.130);
\filldraw[fill=white,draw=abilityOpen,line width=0.65pt] (10.712,1.054) rectangle (10.832,1.174);
\node[rotate=90,text=abilityText] at (0.12,2.04) {Ability pass@3 (\%)};
\end{tikzpicture}

%% file: sections/06_conclusion.tex
\section{Conclusion}
HumanSightBench treats human solvability as an item-level inclusion criterion rather than an aggregate baseline. Its 2,876 questions span seven perceptual categories, with checks for clarity, answer uniqueness, visual grounding, human solvability, and text-only shortcuts. Across eight MLLMs, the strongest model reaches 65.6\% pass@3, while ability-level results expose persistent and uneven bottlenecks. These findings support targeted evaluation of human-easy visual tasks; answer accuracy alone does not certify evidence localization or isolate the cause of an error.

\section*{Ethics Statement}
We collected benchmark data from sources that permit research use or from
internally approved sources. The released benchmark excludes private personal
information, unsafe content, and images for which redistribution rights are
unclear. When images contain identifiable individuals, we follow applicable
consent, anonymization, and review procedures. HumanSightBench is intended for
evaluating visual perception in MLLMs and should not be used as a standalone
indicator of safety, fairness, or real-world deployment readiness.

\section*{Use of AI Assistants}
AI writing assistants were used to help draft and edit the manuscript. All
scientific claims, dataset construction decisions, experimental results, and
final text were reviewed and approved by the authors.

%% file: sections/appendix/a_taxonomy.tex
\section{Full Taxonomy and Ability-Level Coverage}
\label{app:taxonomy}

Figure~\ref{fig:l1-distribution} summarizes the L1 distribution in the main
text. Table~\ref{tab:ability-counts} expands this distribution into all 21 L2
ability groups, reporting their L1--L2 mappings and exact sample counts. The L2
counts sum to the corresponding L1 totals and to 2,876 items overall. The two
groups named Basic Recognition belong to different L1 categories and are
therefore reported separately.

\input{tables/ability_denominators}

\paragraph{Variation within categories.}
L1 totals can hide substantial variation in L2 coverage. In Semantic
Recognition, Color Perception contains 156 of 201 items, whereas Basic
Recognition, Actions and Interactions, and Semantic Relations contain 10 items
each. Within OCR, group sizes range from 70 items for Text Layout and Structure
Understanding to 484 for Character Recognition. These differences motivate
reporting ability-level results alongside category-level aggregates.
Appendix~\ref{app:additional-results} further examines how group sizes and
alternative weighting affect score interpretation.

%% file: tables/ability_denominators.tex
\begin{table}[htbp]
\centering
\caption{Question counts for the 21 evaluated ability groups. IDs identify an L1--L2 path, so the two Basic Recognition groups remain distinct.}
\label{tab:ability-counts}
\small
\setlength{\tabcolsep}{3pt}
\renewcommand{\arraystretch}{1.05}
\begin{tabular}{@{}lp{0.23\linewidth}p{0.57\linewidth}r@{}}
\toprule
ID & L1 category & L2 ability & Items \\
\midrule
1.1 & Grounding & Abstract Concept Grounding & 33 \\
1.2 &  & Counting & 131 \\
\addlinespace[2pt]
2.1 & Semantic Recognition & Basic Recognition & 10 \\
2.2 &  & Actions and Interactions & 10 \\
2.3 &  & States and Attributes & 15 \\
2.4 &  & Color Perception & 156 \\
2.5 &  & Semantic Relations & 10 \\
\addlinespace[2pt]
3.1 & Fine-grained Discr. & Basic Recognition & 100 \\
3.2 &  & Grid Recognition & 81 \\
\addlinespace[2pt]
4.1 & Visual Tracking & Line Tracing & 25 \\
4.2 &  & Contour Recognition & 61 \\
4.3 &  & Visual Reasoning & 32 \\
\addlinespace[2pt]
5.1 & Space Understanding & Relative Orientation and Reference Frame Transformation & 108 \\
5.2 &  & Mirror Image Perception & 64 \\
\addlinespace[2pt]
6.1 & OCR & Text Region Localization and Understanding & 327 \\
6.2 &  & Text Layout and Structure Understanding & 70 \\
6.3 &  & Text Content Recognition & 137 \\
6.4 &  & Character Recognition & 484 \\
6.5 &  & Text Recognition in Less Common Languages & 300 \\
\addlinespace[2pt]
7.1 & Structured Visual Perc. & Shape Recognition & 247 \\
7.2 &  & Chart Recognition & 475 \\
\midrule
& & Total & 2,876 \\
\bottomrule
\end{tabular}
\end{table}

%% file: sections/appendix/b_protocol.tex
\section{Protocol Details}
\label{app:protocol}

\subsection{Human-Level Inclusion Protocol}

\paragraph{Annotation and review groups.}
Candidate question--answer pairs are first constructed by annotator group A
following the predefined ability taxonomy and collection guidelines. Each
candidate is then independently reviewed by annotator group B, which consists
of three ordinary adult annotators. The reviewers are not shown model
predictions and independently answer the same question that is later used for
model evaluation. Items that fail this review are revised and reviewed again
or discarded, following the feedback loop shown in the data construction
pipeline.

\paragraph{Task difficulty and human solvability.}
HumanSightBench targets visual tasks that require no more than
high-school-level knowledge. High-school solvability describes the intended
task difficulty, while actual human solvability is verified by the adult
reviewers in group B. Each candidate must be answerable through ordinary visual
inspection without specialized background knowledge.

\paragraph{Viewing rules.}
Reviewers may inspect the image at its original resolution and may zoom or pan
inside the annotation interface. This rule is fixed because HumanSightBench
tests whether the answer is present in the visual input, rather than whether
the evidence is legible in a particular browser thumbnail. Reviewers may not
crop, enhance, edit, or otherwise transform the image. For all items, the same
image and question are used for human inclusion checks and model evaluation.

\paragraph{Time limit.}
Each item is designed to be solved with ordinary visual inspection rather than
extended search. We therefore impose a per-item time limit of 60 seconds. Items
that require repeated re-reading, extensive manual counting, or unusually long
inspection are flagged as overly time-consuming and are either simplified,
rewritten, or removed.

\paragraph{Inclusion threshold.}
A candidate passes the human-solvability gate only if all three reviewers in
group B independently produce the verified answer after normalization and none
flags the item as ambiguous, impossible to answer from the image, or dependent
on specialized background knowledge. Normalization covers only surface-form
variants such as whitespace, capitalization, punctuation, and equivalent
numeric forms. This threshold is used for item inclusion only; we do not report
the resulting human responses as an aggregate human score.

\paragraph{Disagreement handling.}
If the reviewers disagree, the item is returned for error diagnosis.
Disagreements are assigned to one of four causes: ambiguous wording, multiple
plausible answers, insufficient or unclear visual evidence, or human error on
an otherwise valid item. Items in the first three cases are revised and
reviewed again by independent reviewers; if the revised item fails again, it
is discarded. Items attributed to isolated human errors are retained only
after an additional independent review confirms that the question, answer, and
supporting visual evidence are unambiguous.

\paragraph{Revision and discard criteria.}
An item is revised when the image is valid but the question underspecifies the
target object, counting rule, spatial relation, output format, or acceptable
answer granularity. An item is discarded when the answer cannot be uniquely
verified from the image, when the required evidence is too small or degraded
for ordinary inspection, when the item relies on external knowledge, when it
contains privacy or copyright risk, or when it remains ambiguous after one
revision cycle.

\subsection{Shortcut and Saturation Checks}

\paragraph{Text-only shortcut filtering.}
To reduce language-only solvability, each candidate question is checked without
the image. Items are removed or rewritten when the answer can be inferred from
the question wording, common priors, dataset artifacts, or surrounding text
alone. This filtering is applied before model evaluation so that high model
accuracy cannot be explained by ignoring the image.

\paragraph{Model screening.}
If an item survives the five inclusion gates, model screening removes saturated
items from the candidate set. This ensures that HumanSightBench remains
challenging. The screening step does not define human-level validity: the final
candidate set is conditional on the models and criterion used for screening.
The screening model identities, saturation threshold, and number of items
screened out are not specified in the present manuscript, so overlap between
screening and evaluated models and its impact on rankings is unknown.

\subsection{Model Evaluation Protocol}
\label{app:evaluation-protocol}

\paragraph{Prompt format.}
Each model receives a single image and a short-answer question. The prompt asks
the model to answer concisely and to avoid explanations unless the question
explicitly requests them. For numeric, categorical, or grounding-style
answers, the prompt specifies the expected answer format.

\paragraph{Trials and metric.}
For each model and item, the evaluation uses the first three trials under the
same prompt and image. If $s_{i,r}\in\{0,1\}$ is the judged correctness of
trial $r$ for item $i$, its pass@3 indicator is $\max_{r=1,2,3}s_{i,r}$.
For an evaluation subset $S$, the reported score is
\[
\operatorname{pass@3}(S)=\frac{100}{|S|}\sum_{i\in S}\max_{r=1,2,3}s_{i,r}.
\]
The overall score uses all 2,876 items; category and ability scores restrict
$S$ to the corresponding items. This is an any-correct criterion over three
trials, not the average accuracy of the three answers. Decoding settings are
model-specific and should be documented with the released evaluation metadata.

\paragraph{Scoring.}
DeepSeek-V3.2 (thinking mode) compares the final answer of each response to the verified reference and assigns a binary score. Minor formatting differences and equivalent surface forms do not alter the score. The judge evaluates the final answer not the explanation. A shared failure is when an item is incorrect for every evaluated model in all three trials. Shared-failure rates within abilities are calculated as a ratio of shared-failure items to the number of items in that ability, not to the full size of the benchmark. The scoring rule is mentioned in the manuscript, but neither the complete judge prompt nor a manual audit of the judges' decisions are provided.

%% file: sections/appendix/c_results.tex
\section{Additional Results}
\label{app:additional-results}

This appendix explains how the group sizes in Appendix~\ref{app:taxonomy} affect score interpretation and compares alternative weightings of Table~\ref{tab:main_results}.

\paragraph{Sensitivity to group size.}
Table~\ref{tab:ability-counts} provides the denominator for each ability-level
score. One additional solved item changes pass@3 by $100/n$ percentage points
for a group of $n$ items: 10 points for a 10-item group, compared with about
0.21 points for the 484-item Character Recognition group. Small differences
in reported scores therefore have different granularity across groups.
These increments describe score resolution, not statistical uncertainty;
group-level rankings alone do not establish a reliable ordering of models
beyond the retained items.

\par\vspace{3.25ex plus 1ex minus .2ex}
\noindent
\begin{minipage}[t]{0.46\linewidth}
\vspace{0pt}
{\bfseries Alternative weighting.}\quad Table~\ref{tab:macro-summary} gives the arithmetic average of the seven L1 scores, and the 21 L2 scores separately (both are equal-weighted). These means use the one-decimal percentages displayed in Table~\ref{tab:main_results}; the micro column reproduces its overall score. These are not extra model evaluations but summaries of the results. They are not as heavily weighted towards the large OCR and Structured Visual Perception categories as the micro average is, but L2 averaging gives equal weighting to small ability groups. Neither weighting corrects model-based item selection or provides uncertainty estimates.
\end{minipage}\hfill
\begin{minipage}[t]{0.5\linewidth}
\vspace{0pt}
\centering
\input{tables/macro_summary}
\end{minipage}
\par

%% file: tables/macro_summary.tex
\captionof{table}{Alternative weighting of the same main-table results (\%). Micro reproduces Table~\ref{tab:main_results}; macro columns average its displayed category or ability scores, with equal weight for each group.}
\label{tab:macro-summary}
\small
\setlength{\tabcolsep}{4pt}
\begin{tabular}{@{}lrrr@{}}
\toprule
Model & Micro & L1 macro & L2 macro \\
\midrule
GPT-5.6-Sol & 37.2 & 35.1 & 36.5 \\
GPT-6-Astra & 65.6 & 64.9 & 67.2 \\
GLM-5.3-Flash & 46.2 & 43.4 & 44.7 \\
Seed-Evolving & 58.4 & 56.1 & 57.4 \\
Qwen-3.8-Max & 63.9 & 60.1 & 64.9 \\
GLM-4.6V & 24.0 & 21.5 & 25.2 \\
Kimi K2.6 & 42.2 & 36.4 & 38.9 \\
Qwen3VL-235B-Thinking & 32.4 & 26.2 & 30.0 \\
\bottomrule
\end{tabular}